\documentclass[conference,a4paper]{IEEEtran}
\IEEEoverridecommandlockouts

\usepackage{amsmath}
\usepackage{amsfonts}
\usepackage{amssymb}
\usepackage{cite}
\usepackage{algorithmic}
\usepackage{graphicx}
\usepackage{textcomp}
\usepackage{xcolor}
\usepackage{booktabs}
\usepackage{dsfont}

\def\BibTeX{{\rm B\kern-.05em{\sc i\kern-.025em b}\kern-.08em
    T\kern-.1667em\lower.7ex\hbox{E}\kern-.125emX}}

\begin{document}
\title{\LARGE Joint Model Pruning and Resource Allocation for \\ Wireless Time-triggered Federated Learning}

\author{
    \IEEEauthorblockN{Xinlu Zhang, Yansha Deng, Toktam Mahmoodi}
    \IEEEauthorblockA{Department of Engineering, King's College London, London, UK \\
    \{xinlu.zhang, yansha.deng, toktam.mahmoodi\}@kcl.ac.uk}
}

\maketitle

\begin{abstract}
Time-triggered federated learning, in contrast to conventional event-based federated learning, organizes users into tiers based on fixed time intervals. However, this network still faces challenges due to a growing number of devices and limited wireless bandwidth, increasing issues like stragglers and communication overhead. In this paper, we apply model pruning to wireless Time-triggered systems and jointly study the problem of optimizing the pruning ratio and bandwidth allocation to minimize training loss under communication latency constraints. To solve this joint optimization problem, we perform a convergence analysis on the gradient $l_2$-norm of the asynchronous multi-tier federated learning (FL) model with adaptive model pruning. The convergence upper bound is derived and a joint optimization problem of pruning ratio and wireless bandwidth is defined to minimize the model training loss under a given communication latency constraint. The closed-form solutions for wireless bandwidth and pruning ratio by using KKT conditions are then formulated. As indicated in the simulation experiments, our proposed TT-Prune demonstrates a 40\% reduction in communication cost, compared with the asynchronous multi-tier FL without model pruning, while maintaining the model convergence at the same level.
\end{abstract}


\section{Introduction}
In recent years, wireless federated learning has offered a promising alternative for privacy-preserving distributed machine learning \cite{sun2014data}, which has made great achievements in Big Data Artificial Intelligence. FL avoids transferring user data and only uploads model parameters for each user in each iteration. Based on the manner in which aggregation updates are performed, FL naturally falls into two main categories: Asynchronous(Async) FL and Synchronous(Sync) FL.

Sync FL is widely studied in wireless networks, considering its implementation simplicity and lower communication costs compared to Async FL. The model pruning problem focuses on optimizing aspects such as computation and communication time\cite{jiang2023computation}. Considering the need to maintain or even improve model accuracy, the authors studied the joint user selection and model pruning problems in \cite{liu2021joint}. Since global aggregation can only be triggered after receiving updates from all users in Sync FL, Sync multi-tier networks have been proposed to solve the straggler issue\cite{liu2023adaptive}, but slow users still impact convergence speed and communication efficiency. Async FL provides a more flexible solution by allowing asynchronous aggregation \cite{xie2019asynchronous}. However, Async FL presents new challenges compared to Sync FL, such as uneven training effects, inconsistency in model states, etc \cite{yang2019scheduling}. Despite studies such as weighted averaging and adjusting update frequencies to reduce model inconsistency, the deployment of Async FL is still in its preliminary stages, particularly within wireless networks. 

Several studies have adopted Async FL architecture to tackle the straggler issue. FedAT combines synchronous intra-tier aggregation with asynchronous inter-tier aggregation to mitigate latency and communication hurdles \cite{chai2021fedat}. However, it still faces challenges, like maintaining model accuracy in the case of highly non-IID data and overlooking the impact of wireless communication overhead from the absence of broadcasting. TT-Fed addresses the above issues by conducting global aggregation at fixed intervals, organizing users into tiers based on update times \cite{zhou2022time}. This approach encompasses both Async and Sync FL and falls in the category of asynchronous multi-tier FL when time intervals are fixed.

While time-triggered FL has improved communication overhead compared to traditional FL, the demand for large-scale data still creates challenges for communication costs. Particularly when most devices update rapidly, frequent uploads may cause communication bottlenecks. To address this, studies exploited how to minimize global communication rounds \cite{mcmahan2017communication} or reduce transmitted information \cite{jiang2022model}. However, global round reduction may increase computational costs and reduce model efficiency. 

Model pruning is a method used to reduce the size of FL communication information by simplifying neural networks. Compared to model compression methods, model pruning has low computational complexity and reduces the size of the model without affecting its performance. To improve communication and computation efficiency, PruneFL is proposed as an adaptive model pruning approach. It involves initial pruning at the selected user and further pruning during FL process\cite{jiang2022model}. 
The integration of model pruning introduces new challenges for TT-Fed networks. Larger pruning ratios may lead to changes in user layer assignments, potentially altering the model status of each layer and consequently impacting overall model performance.

Motivated by the above, we propose integrating model pruning and bandwidth allocation in wireless TT-Fed. Our optimization algorithm combines resource allocation and pruning, balancing communication efficiency and convergence speed by jointly optimizing the pruning ratio and bandwidth allocation.
The contributions of our research can be summarized as follows:
\begin{enumerate}
\item{We develop a joint optimization framework called TT-Prune for TT-Fed for resource-constrained and large-scale information transfer wireless networks. We formulate a joint model pruning and bandwidth problem to maintain model accuracy.}

\item{We derive mathematical analytical formulas for convergence speed and learning latency. According to the analytical results, a closed-form solution for the optimal pruning ratio and bandwidth allocation is derived. We decouple the problem into tractable sub-problems and deploy KKT conditions.}

\item{We propose setting an identical pruning ratio within each tier to minimize the extra computation further. We evaluate the performance of TT-Prune in comparison with the other three strategies. Simulation results show that our TT-Prune reduces the communication cost by about 40\% compared to no pruning and achieves the same level of learning accuracy, proving the effectiveness of our proposed TT-Prune.}
\end{enumerate}

The rest of the paper is organized as follows: Section II presents the system model. Section III covers convergence analysis and optimal solutions. Section IV outlines the simulation results. Finally, Section V concludes the paper.

\section{System Model and problem formulation}

In this section, we begin by outlining the Async multi-tier TT-Fed framework, followed by a brief description of model pruning. Next, the communication and computation latency are explained. Finally, the problem formation associated with our TT-Prune is presented.
\subsection {Asynchronous multi-tier FL Framework}
Consider a multi-tier wireless network consisting of an edge server, denoted as $\mathcal{K}$, a set of $\mathcal{U}$ users, denoted as $\mathcal{U}  = \{  u=1,2,\ldots,U \}$. Each user $u \in \mathcal{U} $ possesses a local data set $D_u$, with a data volume $|D_u|$. Each data sample is represented by a set of input-output pairs $(X_{u,i}, Y_{u,i}) , i\in D_u $ , where $X_{u,i}$ denotes the $i$-th input data and $Y_{u,i}$ represents the corresponding ground truth for $X_{u,i}$. The entire network's data volume is $D$, where $D = \sum_{u \in U} D_u
$. $\textbf{\textit{w}}$ represent the network model and the loss function is denoted by  $ \ell(x_i,y_i|\textbf{\textit{w}})$.

When adopting the asynchronous multi-tier Federated Learning model, we use the TT-Fed as the benchmark model \cite{zhou2022time}. In this system, participants are naturally classified into different levels according to fixed global aggregation round period $\Delta T $. We assume the time for the slowest user to complete one local training round is $T$. Then, the number of tiers is $M = \lceil T/\Delta T \rceil$ tiers, where $\lceil \cdot \rceil$ is the ceiling function. 

\begin{figure*}[t!]
  \centering
  \includegraphics[width=17cm]{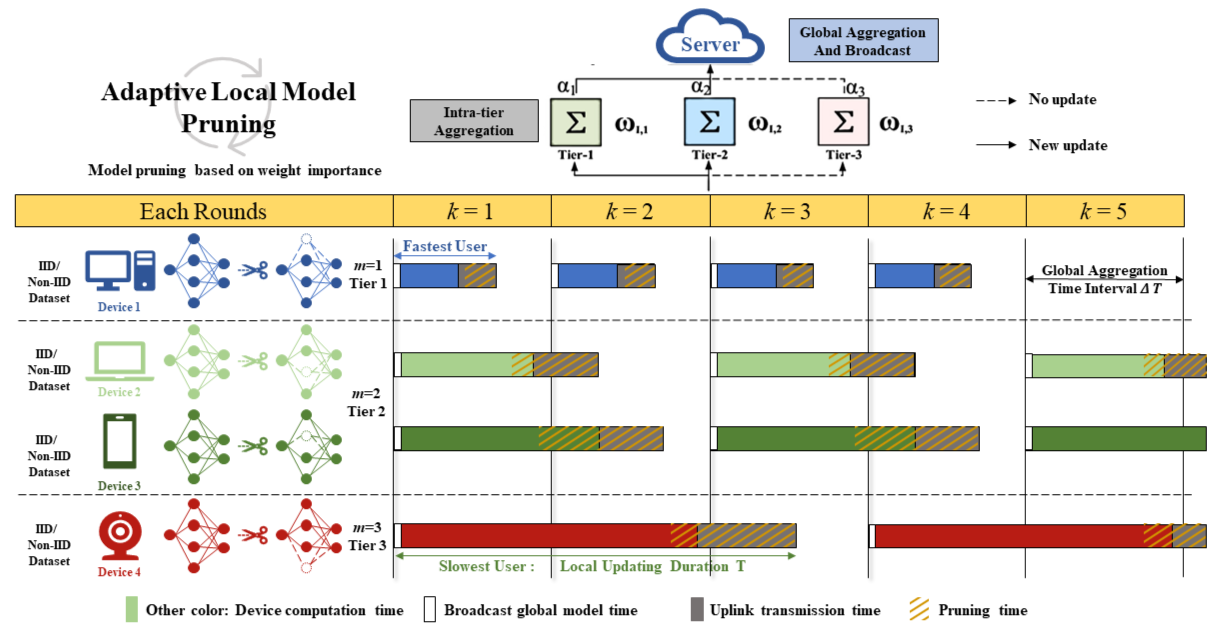}
  \caption{The work-flow of Pruned TT-Fed under the given aggregation duration $\Delta T $.}
  \vspace{-5pt}
  \label{TTFL}
\end{figure*}

The specific aggregation process used in Asynchronous multi-tier FL is shown in Fig.\ref{TTFL}. In a scenario with four participants, they are divided into three tiers based on $\Delta T$. It is important to note that users in the first tier update locally in each global aggregation, while those in the second tier update every two rounds, and so forth. The variable $k$ represents the $k$-th round of model aggregation, and $m$ represents the $m$-th tier. To decide if the $m$-th tier joins the current aggregation, we check if $k$  \textbf{mod} $m$ is equal to zero. For example, the user 4 don't update in the 1st and 2nd global aggregation since the $ k $  \textbf{mod} $m$ $\neq 0$ ( $k=1$ or 2 and $m = 3$).

The optimization goal of Async multi-tier FL is the same as traditional FL, that is, to minimize the empirical loss $\mathcal{L}(\mathbf{w})$ over all distributed training data. It can be formulated as
\begin{equation}
\min _{\mathbf{w}} \mathcal{L}(\mathbf{w})=\frac{ \sum_{u \in \mathcal{U}}\sum_{i \in \mathcal{D}_u} \ell\left(\mathbf{x}_{u,i}, \mathbf{y}_{u,i} \mid \mathbf{w}\right)}{|\mathcal{D}|}.
\label{eq1}
\end{equation}

The iterative learning process, which can be partitioned into four steps, is introduced as follows:
\subsubsection {Global Model Broadcast}
At the end of the ($k$-1)-th round of global aggregation, the server broadcasts the latest global model to the users who participated in the ($k$-1)-th round of aggregation. Users participating in the aggregation can be represented by \{$S_m$ $|k-1$ mod m = 0, $\forall m \in M $\}. 
\subsubsection {Local Model Learning}
The $k$-th global aggregation is provided by users who belong to the $m$-th tier and are trained on the global model $w_{G}^{k-m}$. The generated local model that is ready to be uploaded can be represented as
\begin{equation}
w_{L,u}^{k} = w_G^{k-m} - \lambda \mathbb{E}\left[\nabla \ell(w_G^{k-m} ; x_{u,i}, y_{u,i})\right], \forall u \in S_m ,
\label{eq2}
\end{equation}
where $ \lambda $ is the learning rate and the $w_G^{k-m}$ is the global model that user $u$ received at the $(k-m)$-th global round. 
\subsubsection {Intra-tier Aggregation }
 Intra-tier aggregation refers to combining local models within the same tier before global aggregation. In the $k$-th round, the intra-tier aggregation process of the $m$-th tier is presented as 
\begin{equation}
w_{\mathrm{I}, m}^k=\sum_{u \in \mathcal{S}_m} \frac{D_u w_{\mathrm{L}, u}^k}{\sum_{u \in \mathcal{S}_m} D_u}.
\label{eq3}
\end{equation}
\subsubsection {Global Aggregation }
The aggregation formula of the $k$-th round of global aggregation is presented as
\begin{equation}
\begin{aligned}
    w_{G}^k=&\sum_{m=1}^M \ \mathds{1} \{k \bmod m=0\} \alpha_m^k w_{\mathrm{I}, m}^k \\
    & +\sum_{m=1}^M(1-\ \mathds{1}\{k \bmod m=0\}) \alpha_m^k w_{G}^{k-1},
\label{eq4}
\end{aligned}
\end{equation}
where $\alpha_m^k$ is the aggregation weight of models from the $m$-th tier at the $k$-th global aggregation round, and $\sum_{m=1}^M\alpha_m^k=1$. In Eq. (\ref{eq4}), $\ \mathds{1}\{ \cdot \}$ is the indicator function. The first term of right-hand in Eq. (\ref{eq5}) is the weighted sum of uploaded models, and the second term is the latest global model $w_{\mathrm{G}}^{k-1}$ multiplied by the corresponding aggregation weight.

To balance the computation bias towards fast tiers, we adopt a heuristic weighting scheme (as seen in Eq. (\ref{eq5}), where $\lfloor \cdot \rfloor$ denotes the floor function). Hence, the aggregation weight for models from the $m$-th tier at the $k$-th global aggregation round is given by
\begin{equation}
\alpha_m^k=\frac{\left\lfloor\frac{k}{M+1-m}\right\rfloor}{\sum_{m=1}^M\left\lfloor\frac{k}{m}\right\rfloor}.
\label{eq5}
\end{equation}

By adjusting the global aggregation time $\Delta T$, TT-Fed is transformed into a different type of FL. If $\Delta T$ is set to approach 0, each user is assigned to an independent tier. So TT-Fed transforms into Async FL, where global aggregation is triggered whenever a local update occurs. Conversely, if $\Delta T$ is set to approach infinity, it transforms into Sync FL. All users are categorized into the same tier, where all users must wait for the slowest user before aggregation. TT-Fed represents a special case of Sync and Async FL. When $\Delta T$ is set between the fastest and slowest local user times, Async multi-tier networks emerge. Users with similar training times are grouped together, updating independently without waiting for the slowest user. Thus, we use TT-fed as a benchmark for pruning in the Async multi-tier FL framework.
\subsection {Federated Learning With Model Pruning}
Due to limitations in device computing power and wireless communication capacity, model pruning, as an effective compression technique, aims to retain model performance while removing unimportant weights. Pruned units do not require forward and backward passes or gradient updates. We set the pruning ratio for the $m$-th tier in the $k$-th round of global aggregation as
\begin{equation}
\rho_{k,m}= \frac{M_{p}}{M},
\label{eq6}
\end{equation}
where \(M_{p}\) represents the number of model parameters to be pruned, where \(M\) denotes the original size of the model.

We set the model pruning ratio for each tier instead of each user because models within the same tier share significant similarities. Therefore, the pruning ratio has roughly similar impacts on the latency of models within the same tier. Additionally, calculating the pruning ratio for each device incurs significantly higher computational costs compared to per-tier calculations.

To determine weight importance, we adopt a lower-complexity method, which involves comparing the differences between the weights of each device's local models before and after updating. It simplifies the calculation process while ensuring effectiveness, which can be described as ($ \hat{w}_{u,j}$ means updated weight)
\begin{equation}
 \hat{I}_{u,j} = |w_{u,j} - \hat{w}_{u,j}|.
\label{eq7}
\end{equation}

No additional computation is required when removing unimportant weights by importance measures, as updated weights can be found in the backpropagation. It significantly reduces the processing time during updating while also reducing the model size and easing the communication load. The total number of parameters in the entire model can be simply calculated as
\begin{equation}
W_{u, m}=\left\lceil(1-\rho_{k,m}) W_{u, m, \text { in }}\right\rceil\left\lceil(1-\rho_{k,m}) W_{u, m, \text { out }}\right\rceil,
\label{eq8}
\end{equation}
where $W_{u, m, \text { in }}$ and $W_{u,m, \text { out }}$ correspond to the number of input and output weights, respectively. $\rho_{k,m}$ to denote the pruning ratio of the $u$-th device in the following sections since the pruning ratio is the same at the same tier.

\subsection {Latency In Each Global Iteration}
Global iterations mainly occur during inter-tier and global aggregation, concentrating model updates at the central server and local devices. We explain latency overhead by analyzing the entire global iteration process, which is divided into four main steps:

\subsubsection {Local Model Update With Pruning} When the $u$-th $m$ tier local model receives the model parameters from the $k$-th round of global aggregation, it prunes the model using a pruning mask (derived from the weight importance function we defined as Eq. (\ref{eq6})). The pruned model weights can be represented as
\begin{equation}
\boldsymbol{\widetilde{w}}_{k, m}^{u}=\boldsymbol{w}_{k, m}^{u} \odot \boldsymbol{m}_{k, m}^{u}.
\label{eq9}
\end{equation}

For the pruning mask \resizebox{0.045\textwidth}{!}{$ \boldsymbol{m}_{k, m}^{u}$}, if \resizebox{0.13\textwidth}{!}{ $\boldsymbol{m}_{k, m}^{u,j} = 1, \boldsymbol{\widetilde{w}}_{k, m}^{u}$ }contains the $j$-th model weight, otherwise, $m_{k, m}^{u,j}=0$. 

We primarily emphasize weight pruning in the fully-connected layer over the convolutional layer. This decision is based on the observation that pruning in the convolutional layer often reduces the robustness of the CNN. Given a pruning ratio $\rho_{k,m}$ of the $u$-th mobile device, the number of weights after pruning is calculated as
\begin{equation}
W_{\rho_{k, m}}= W_{u,\text{conv}}+(1-\rho_{k,m})W_{u,\text{fully}}.
\label{eq11}
\end{equation}
where $W_{u, \text{conv}}$ represents the number of weights in the convolutional layer, and $W_{u, \text{fully}}$ denotes the number of weights in the fully connected layer. 

With the change of model weight number, the computation latency for each user device is affected during the updating period. To calculate the computation latency for the $u$-th local device, we assume that the number of CPU cycles required to process a weight of the model is denoted as $c_u$, and the CPU frequency of the $u$-th user is $f_u$. Therefore, the time taken by the $u$-th user to complete local updates during the $k$-th global aggregation round is given by \cite{yang2020energy}
\begin{equation}
\tau_{u, k}^\text{cp}=\zeta \frac{  W_{\rho_{k, m}} c_u}{f_u},
\label{eq12}
\end{equation}
where $\zeta$ represents the number of local training epochs 
\subsubsection {Local Model Uplink Transmission} Assuming the size of the model transmitted to the server as \(Z\) bits. Then, the achievable rate \(R_{k,u}^\text{up}\) and uplink communication time \(\tau_{k,u}^\text{cm}\) for the \(u\)-th user in the \(k\)-th global aggregation round can be determined by
\begin{equation}
\begin{aligned}
R_{k,u}^{\text{up}} =& b_{k,m} B \log_2 \left(1 + \frac{p g_{k,u}}{N_0}\right), \\
Z =  \hat{q}W_{\rho_{k, m} } ,\quad  & \tau_{k,u}^\text{cm} =\frac{Z}{R_{k,u}}, \quad \text{for all } u \in U .
\end{aligned}
\label{eq13}
\end{equation}

The $\hat{q}$ means the quantization bit. The bandwidth \(b_{k,m}\) allocated to the \(m\)-th tier in the \(k\)-th global aggregation round, the uplink transmission power \(p\) (assuming equal transmission power for each user), and the noise power \(N_o\). \(g_{k,u}\) is the channel gain between the $u$-th local device and central serve.

\subsubsection {Global Model Aggregation and Broadcast}The edge server aggregates local models from all tier users as Eq. (\ref{eq4}). Given the server's abundant computational resources, global aggregation at the central server is computationally efficient. Hence, we ignore computation time at the edge server. Additionally, the transmission delay during broadcasting can be neglected due to the higher downlink transmission rate.

\subsection {Problem Formulation}
Based on the aforementioned system model and considered latency, we formulate an optimization problem with respect to maximizing convergence speed. Maximizing convergence speed is equivalent to minimizing the $l_2$-norm of the gradient. This allows for finding the optimal parameter update direction and accelerating convergence within limited training iterations. We aim to maximize convergence speed while ensuring learning latency by limiting latency and optimally allocating pruning ratio and bandwidth. At the end, the optimization problem is described as
\begin{subequations}
\begin{equation}
    \min_{b_{k,m}, \rho_{k, m}} \sum_{k=1}^K \mathbb{E}\left\|\nabla F\left(w_G^k\right)\right\|^2,
    \label{eq15a}
\end{equation}
\begin{equation}
     \left(1-\rho_{k,m}\right) W_{k,m}^u \left(\zeta \frac{c_u}{f_u}+\frac{\hat{q}}{R_{m, u}^k}\right) \leq m \Delta T ,
    \label{eq15b}
\end{equation}
\begin{equation}
    \sum_{m=1}^M b_{k,m} \leq 1,
    \label{eq15c}
\end{equation}
\begin{equation}
    0 \leq b_{k,m} \leq 1,
    \label{eq15d}
\end{equation}
\begin{equation}
    \rho_{k,m} \in[0,1].
    \label{eq15e}
\end{equation}
\label{eq15}
\end{subequations}

Eq. (\ref{eq15a}) minimizes the $l_2$-norm of the gradient for faster convergence. Eq. (\ref{eq15b}) ensures timely learning completion for each user before aggregation. Additionally, Eq. (\ref{eq15c}) and Eq. (\ref{eq15d}) limit allocated bandwidth between [0, 1]. Eq. (\ref{eq15e}) bounds the model pruning ratio within [0, 1]. 

Clearly, The optimization problem in Eq. (\ref{eq15}) constitutes a mixed-integer nonlinear programming problem characterized by its non-convexity and impracticality in obtaining optimal solutions directly. Consequently, we decompose the complex optimization problem into manageable sub-problems.

\section {Convergence Analysis and Optimal Solution}

\subsection {Convergence analysis}
The average $l_2$-norm is utilized to assess the convergence performance \cite{ghadimi2013stochastic}, given that neural networks are typically non-convex. Before conducting the convergence analysis, the following assumptions are made for the analysis:
\begin{enumerate}
\item{ \textbf{The L-smooth characteristic of loss function $F$}, i.e.:}
\begin{equation}
    \left\|\nabla F_n\left(\boldsymbol{w}_1\right)-\nabla F_n\left(\boldsymbol{w}_2\right)\right\| \leq L\left\|\boldsymbol{w}_1-\boldsymbol{w}_2\right\|,\\
     \label{eq16}
\end{equation}
where $L$ is a positive constant and $\left\|\boldsymbol{w}_1-\boldsymbol{w}_2\right\|$ is the norm of 
 $\boldsymbol{w}_1-\boldsymbol{w}_2$. 
\item{ \textbf{Bounded global gradient change within $\boldsymbol{m \in  M}$ training rounds}, i.e.:}
\begin{equation}
\begin{aligned}
\left(w_{\mathrm{G}}^{k-m}-w_{\mathrm{G}}^{k-1}\right)^{\top} \nabla  F\left(w_{\mathrm{G}}^{k-1}\right) & \leq  \delta\left\|\nabla  F\left(w_{\mathrm{G}}^{k-1}\right)\right\|^2, \\
\left\|w_{\mathrm{G}}^{k-m}-w_{\mathrm{G}}^{k-1}\right\| & \leq  \varepsilon,
\end{aligned}
\label{eq18}
\end{equation}
where $\delta$ and $\varepsilon$ are positive constants.
\item{ \textbf{Bounded local gradient change within $\boldsymbol{m\in M}$ training rounds}, i.e.:}
\begin{equation}
\begin{aligned}
& \left\|\nabla f\left(w_{\mathrm{G}}^{k-m}\right)\right\| \leq \beta\left\|\nabla f\left(w_{\mathrm{G}}^{k-1}\right)\right\|, \\
& \left\|\nabla f\left(w_{\mathrm{G}}^{k-1}\right)-\nabla f\left(w_{\mathrm{G}}^{k-m}\right)\right\| \leq \phi,
\end{aligned}
\label{eq19}
\end{equation}
where $\beta$ and $\phi$ are both positive constants. 
\item{ \textbf{Pruning-induced Noise in convergence}, i.e.:}
\begin{equation}
    \mathbb{E}\left\|\boldsymbol{w}_{k, n}^{q, e}-\boldsymbol{w}_{k, n}^{q, e} \odot \boldsymbol{m}_{k, n}^{q, e}\right\|^2 \leq \rho_{n, e} D^2,
    \label{eq20}
\end{equation}
where \textit{D} is a positive constant \cite{stich2018sparsified}.
\end{enumerate}

These common assumptions, met by widely used loss functions like cross-entropy, form the basis for deriving the convergence rate with network pruning, as shown in the following theorem.

\textbf{Theorem 1:} When the above assumptions holds and $\frac{1}{4 L} > \delta$, the upper bound on the average $l_2$-norm of the gradient after $S$ iterations in TT-Fed using pruning can be derived as
\begin{equation}
\begin{aligned}
&\frac{\xi}{4L} \sum_{k=1}^K \mathbb{E}\left\|\nabla F\left(w_G^k\right)\right\|^2 \leq \\
&\frac{\mathbb{E}\left[F\left(w^0\right)\right]-
\left[F\left(w^*\right)\right]}{1-4L\delta} +\xi \frac{3LD^2}{(1-4L\delta)M} \sum_{k=1}^K \sum_{m=1}^M \frac{S_m^2}{D_m^2} \rho_{m, k}\\
& + \frac{K \xi L \varepsilon^2}{1-4L\delta} +\frac{\xi K}{(1-4L\delta)2M} (3L  \Omega_1 +\frac{M-1}{LD^2}  \Omega_2)   .
\label{eq21}
\end{aligned}
\end{equation}

In Eq. (\ref{eq21}), $K$ is the number of global communication rounds, $M$ is the total number of model tiers, $D$ represents the total number of data samples used in the training process, and $S_m$ means the number of users in $m$-th tier. $\xi$ is the parameter at median value theorem, where 
\begin{equation}
    \Omega_1 = \sum_{m=1}^M \frac{S_m^2}{D_m^2} \varepsilon^2  \quad \text{and} \quad
    \Omega_2 = \sum_{m=1}^M \sum_{j=1, j \neq m}^M S_j^2 \phi^2  .
\label{eq22}
\end{equation}

\subsection {Optimal Solution}

\textbf{1). Optimal Pruning Ratio Analysis:}  Once a constant value is assigned to the bandwidth, the problem described in Eq. (\ref{eq21}) can be viewed as a linear programming problem. Since the optimization solution is set the same for each tier, $u$ here represents the average value of users in each tier. Subsequently, we derive the optimal solution for the pruning ratio based on the theorem presented below. 

\textbf{Theorem 2:} Under the given bandwidth allocation, the optimal ratio of the device in the $m$-tier associated with $k$-th global communication round should satisfy
\begin{equation}
     \rho^*_{m, k} \geqslant \left(1-\frac{m \Delta T - W_{u, \text{conv}} \left(\xi \frac{ c_u}{f_u}+\frac{\hat{q}}{R_{m,u}^{\text{k}}}\right)  }{(\xi \frac{ c_u}{f_u}+\frac{\hat{q}}{R_{m,u}^{\text{k}}})W_{u, \text{fully}}}\right)^+ ,
     \label{eq25}
\end{equation}
where the function \((x)^+\) means \(max(x,0)^+\).



\textbf{2). Optimal Bandwidth Allocation:}
Based on the optimal pruning ratio and the upper bound of the average $l_2$-norm of the gradient, the original problem can be rewritten as

\begin{equation}
      \min_{b^*,\rho^*}  \boldsymbol{\Delta} \sum_{k=1}^K \sum_{m=1}^M \frac{S_m^2}{D_m^2}\left(1-\frac{m \Delta T - W_{u, \text{conv}} \left(\xi \frac{ c_u}{f_u}+\frac{\hat{q}}{R_{m,u}^{\text{k}}}\right)  }{(\xi \frac{ c_u}{f_u}+\frac{\hat{q}}{R_{m,u}^{\text{k}}})W_{u, \text{fully}}}\right).
      \label{eq26}
\end{equation}

In order to get the optimal solution for bandwidth, we need to minimize Eq. (\ref{eq26}). where $\boldsymbol{\Delta} =  \frac{3LD^2}{(1-4L\delta)M} $.

\textbf{Lemma 1: The problem Eq. (\ref{eq26}) is convex. }

According to \textbf{Lemma 1}, we perform multiple derivative operations on Eq. (\ref{eq26}). Then, the optimal wireless bandwidth is given by using the Lagrange multiplier method \cite{polik2010interior}.

\textbf{Theorem 3:} Setting $\lambda^*$ as the optimal Lagrange multiplier, to achieve optimal FL performance, the optimal bandwidth allocated to the $u$-th device in the $m$-th tier should satisfy

\begin{equation}
b_{k,m}^*=\frac{f_u\left(\sqrt{\frac{\boldsymbol{\Delta}S_m^2 m\Delta T\hat{q}W_{u, \text{fully}} B  \log _2\left(1+\frac{p g_{k,u}}{N_0}\right)}{D_m^2 \lambda^*}}-\hat{q}W_{u, \text{fully}}\right)}{BW_{u, \text{fully}}\xi c_u  \log _2\left(1+\frac{p g_{k,u}}{N_0}\right)}.
\label{eq27}
\end{equation}

To dynamically adapt to the changing wireless network environment over time, the optimized algorithm needs to be executed once during each global aggregation.

Due to space limitations, all the proof is provided in the journal version.

This paper mentions several baselines to reduce the communication volume of wireless TT-Fed in addition to the proposed TT-Prune. 
All schemes are described as follows:
\begin{enumerate}
    \item \textbf{TT-Prune}: Both the pruning ratio and bandwidth allocation are optimized according to Theorem 3. 
    \item \textbf{Equal Bandwidth}: The pruning ratio is optimized according to Theorem 3 while the bandwidth is equally allocated to all users. 
    \item \textbf{Equal Pruning}: The pruning ratio is the same for each tier, and the bandwidth is equally allocated to all users. 
    \item \textbf{No Pruning}: The bandwidth is equally allocated to all users, and model pruning is not implemented.
\end{enumerate}

\section {Numerical Results}
In the simulations, we consider a circular network area with a radius $ r = 500 m $ and one BS, inside which $U$ = 12 users are uniformly distributed. Each user trains a traditional CNN model for image classification using the cross-entry loss function. We employ the MNIST and Fashion MNIST (FMNIST) to validate TT-Prune's effectiveness. The dataset has both IID and Non-iid characteristics. 
The transmission power of devices is 28 dBm, CPU frequency is about 1$\sim$ 5 GHz. AWGN noise power is -174 dBm/Hz, the quantization bit is 64, local epochs are 5, global communication rounds are 10, the learning rate is 0.0015, the channel bandwidth is 20 MHz, and time interval $\Delta T = 0.7$ \textcolor{red}{T}. This simulation evaluates the performance of TT-Prune by comparing its test results with other considered optimization schemes.

\begin{figure}[htbp]
\centering
\includegraphics[width=7cm]{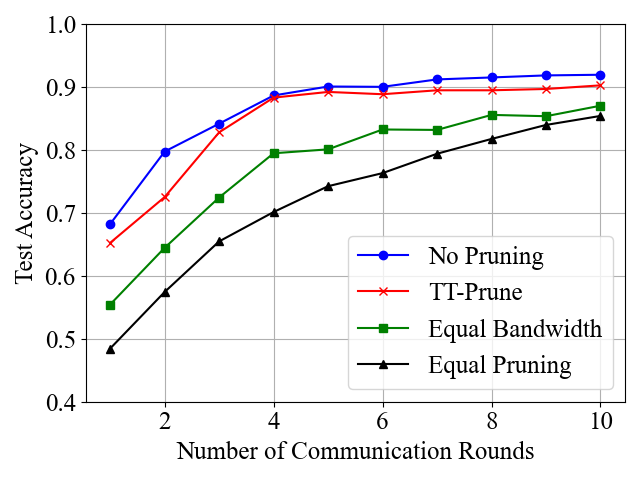}
\caption{Test accuracy of TT-Fed under different scheme in Non-IID FMNIST dataset}
 \vspace{-5pt}
\label{fig2}
\end{figure}

\begin{figure}[htbp]
\centering
\includegraphics[width=7cm]{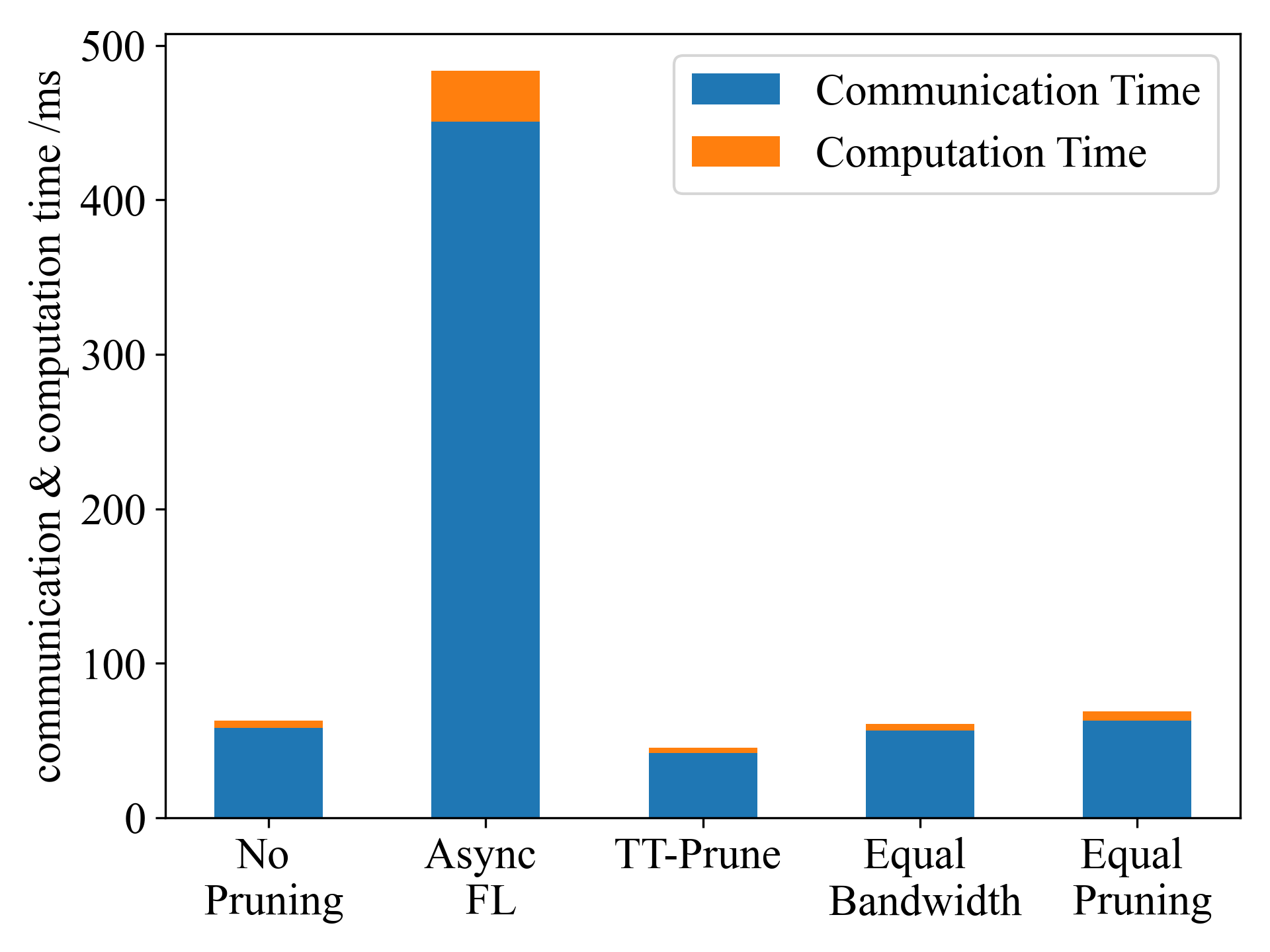}
\caption{Performance required for TT-Prune and other schemes to achieve 80\% accuracy in Non-IID FMNIST dataset }
\vspace{-7pt}
\label{fig3}
\end{figure}

Considering the characteristics of TT-Fed, the number of tiers in the network dynamically changes based on the provided latency budgets. The choice of latency budget impacts the model's convergence rate. To better demonstrate the benefits of the proposed TT-Prune on the model, We set a two-tier TT-Fed as the pruning target. When employing the TT-Prune scheme, if the model only conducts first-tier device aggregation, it changes to a single-tier aggregation state. It cannot benefit from multi-tier optimization strategies in this state. Consequently, optimization is performed individually for each user.

Fig.\ref{fig2} shows the test accuracy of four TT-Fed schemes under Non-iid dataset, which reveals the performance gap under different pruning ratios and bandwidth allocation policies. It can be seen that there are large differences between the schemes, which may be the result of a small amount of missing data or bias in each user. We see a close gap between TT-Prune and the non-pruning scheme, which confirms its effectiveness. For the equal bandwidth scheme can be seen to be lower than that of the proposed TT-Prune. This is because the important weight may be pruned during the learning process. Equal pruning, as it directly sets the pruning ratio without considering the model's convergence speed, results in the worst performance among the examined methods.

Fig.\ref{fig3} plots the time required to reach 80\% accuracy for the five schemes. Async FL is used to highlight the communication efficiency of these schemes \cite{xie2019asynchronous}, and global aggregation is triggered when the server receives an update from any user. we can observe that the communication time is significantly greater than the computation time. Compared to no pruning TT-Fed, TT-Prune reduces latency by approximately 40\%. The reason for this reduction is that the algorithm prunes less important weights based on channel conditions, thereby reducing latency. Although Equal pruning greatly reduces the size of the transmission model because it deletes some important weights, the model requires more rounds of communication, which means more time, to achieve the required accuracy. Additionally, it is evident that compared to Async FL, TT-Prune significantly improves model communication time, decreasing it to approximately one-tenth.

\section{Conclusion}

This paper proposed a joint optimization framework for model pruning and bandwidth allocation in wireless TT-Fed to enhance communication and learning efficiency. Model pruning is utilized to reduce model size dynamically based on network conditions. We first conduct an analysis of the convergence rate in TT-Fed with model pruning to evaluate its performance. Subsequently, we utilized KKT conditions to jointly optimize the pruning ratio and wireless resource allocation under latency and bandwidth constraints. Our simulation results show that TT-Prune achieves a 40\% reduction in total latency compared to models without pruning while maintaining the same level of learning accuracy. 



\footnotesize
\bibliographystyle{ieeetr}
\bibliography{conference_main}

\vspace{12pt}
\color{red}

\end{document}